\documentclass{article}

\usepackage{arxiv}

\usepackage{orcidlink}

\usepackage[utf8]{inputenc} % allow utf-8 input
\usepackage[T1]{fontenc}    % use 8-bit T1 fonts
\usepackage{hyperref}       % hyperlinks
\usepackage{url}            % simple URL typesetting
\usepackage{booktabs}       % professional-quality tables
\usepackage{amsfonts}       % blackboard math symbols
\usepackage{nicefrac}       % compact symbols for 1/2, etc.
\usepackage{microtype}      % microtypography
\usepackage{lipsum}
\usepackage{graphicx}
\usepackage{authblk}
\graphicspath{ {./images/} }

\title{Mapping the Web of Science, a large-scale graph and text-based dataset with llm embeddings}

\author[1]{Tim Kunt \orcidlink{0009-0006-5732-3208}}
\author[1]{Annika Buchholz \orcidlink{0009-0000-2110-6929}}
\author[1]{Imene Khebouri \orcidlink{0000-0002-2389-8795}}
\author[2,3]{Thorsten Koch \orcidlink{0000-0002-1967-0077}}
\author[1]{Ida Litzel \orcidlink{0009-0000-0166-7809}}
\author[1,4]{Thi Huong Vu \orcidlink{0009-0007-4869-0505}}

\affil[1]{Digital Data and Information for Society, Science, and Culture,  \protect\\ Zuse Institute Berlin, 
Takustr. 7, 14195 Berlin, Germany}

\affil[2]{Software and Algorithms for Discrete Optimization, \protect\\Technische Universität Berlin,
Straße des 17. Juni 135, 10623 Berlin, Germany}

\affil[3]{Applied Optimization, Zuse Institute Berlin,\\
Takustr. 7, 14195 Berlin, Germany}

\affil[4]{Institute of Mathematics, Vietnam Academy of Science and Technology,
10072 Hanoi, Vietnam}

\begin{document}
\maketitle
\begin{abstract}
Large text data sets, such as publications, websites, and other text-based media, inherit two distinct types of features: (1) the text itself, its information conveyed through semantics, and (2) its relationship to other texts through links, references, or shared attributes. While the latter can be described as a graph structure and can be handled by a range of established algorithms for classification and prediction, the former has recently gained new potential through the use of LLM embedding models.
Demonstrating these possibilities and their practicability, we investigate the Web of Science dataset, containing ~56 million scientific publications through the lens of our proposed embedding method, revealing a self-structured landscape of texts.
\end{abstract}

\section{Introduction}
\label{sec:1}
\subsection{Towards a Map of Sciences}
A primary question of scientometrics is the categorization of sciences into subjects. This categorization and systems of knowledge organization are often affirmative and never neutral \cite{rusch-map}. 
As part of the ongoing project FAN \cite{fan}, our approach combines graph-theoretical methods and text processing on large-scale citation graphs, aiming to develop bottom-up and robust methods. From a constructivist perspective, graph theory is a meaningful and elegant abstraction of the relations formed between (text-based) media through references, citations, and shared features of their metadata. Natural language processing (NLP) methods derive an analysis of individual texts. Combining these two approaches promises to compensate for their respective shortcomings, as they are independent of each other in large parts. However, it also comes with a series of difficulties, as mathematically the two employ methods that are not immediately compatible, and an attempt at translation may result in a substantial loss of information. \\ 
Existing algorithms that map such databases based on the citation graph are the current standard and are suitable for clustering and classifying the dataset to a reasonably fine degree. They leverage the relatedness of individual records based on citations and co-citations \cite{Waltman}. A comprehensive study and comparison are given by \cite{clustering}. Web of Science \cite{web-of-science} itself introduced a classification schema \cite{citationtopics} based on the Leiden algorithm \cite{Traag}, although the details of its implementation are not publicly documented. \\
On the other hand, text representation methods rapidly evolved over the last years \cite{Siebers}, which motivates reexamining their suitability for clustering and classification \cite{Keraghel}\cite{Petukhova}.
This study primarily demonstrates a text–embedding–based approach. The developed methods lay out and examine a greater map of sciences on the basis of Web of Science. 
Further, we briefly examine how this approach compares with graph-based methods, motivating the development of a robust hybrid method.

\subsection{Text Embeddings via LLMs}
LLM-based embeddings, mapping any text onto an element of a high-dimensional vector space, are the latest addition in the toolbox of text representation techniques \cite{Petukhova}. Trained on huge datasets covering a large variety of language, they appear to be a powerful one-fits-all solution to the search for a mapping that is not overly complex in its output, but still retains the distinction between even finer semantic differences. We may, however, find that the performance of generic embeddings is limited and the aforementioned generality comes at the cost of precision when applied to domain-specific datasets, \cite{sarma2018}. A limitation, tailored or fine-tuned NLP methods may be better equipped to overcome.\\
For the purpose of representing the abstracts of scientific publications, we argue that LLM embeddings are potentially a suitable choice, and that many of the associated drawbacks do not necessarily apply to the text form of scientific abstracts, as they:
\begin{itemize}
    \item  are \textbf{descriptive}, summarizing purpose, scope, and conclusion of a paper, and contain the relevant \textbf{keywords} on their topic, guiding embedding (and other NLP) methods
    \item  usually don't employ linguistic devices or other \textbf{semantic difficulties} for LLMs
    \item  are overall \textbf{homogeneous}, almost formulaic text-form, advantageous for large-scale text processing and representation
\end{itemize}
In conclusion, an LLM embedding of abstracts or full-texts of scientific articles should capture the broader semantic differences of its texts and thereby create a meaningful interpretable map of the entire dataset in the embedding space - that is the premise of this work. \\
Going from there, we investigate whether this approach is suitable as a foundation for the prediction and classification of a text's location in said landscape, which in the context of scientific publications could be interpreted as subject or topic. The resulting map was then compared to the existing taxonomy of subjects and sciences to verify and support the findings, as well as to investigate proximity and cluster formations within.

\section{Method} 
\subsection{Embedding}
For LLM embedding, we use the \textit{ollama} framework and run all computations with both the embedding models \textit{mxbai-embed-large:335m} \cite{mxbai} and \textit{nomic-embed-text:v1.5} \cite{nomic}. Different embedding models produce different representations of a dataset in their embedding space, and consequently, differing categorizations or clusterings computed within those spaces, as discussed by \cite{Petukhova}. Given our focus is on the method itself, we limited ourselves to two LLM embedding models: \textit{mxbai-embed-large:335m} with 1024 dimensions and \textit{nomic-embed-text:v1.5} with 768 dimensions as a reproducible open source and open data model with a comparable competitive performance on our dataset.\\
\\
We return the normalized vectors and consider cosine similarity from here on as a distance metric. 
Beautifully, the resulting point cloud can be thought of as forming continents on a globe, as they all lie on an $n$-sphere (where $n$ is the dimension of the embedding model). A clustering or categorization of those points can be thought of as drawing a map on this globe (Fig. \ref{fig:3}).

\subsection{Centers, Clusters and Distances}
Building on the embedding vectors and the citation graph, a wider range of analytical methods can be applied, a selection of which is outlined in the following.\\
Firstly, we compute the center-points of each of the 255 unique subjects in Web of Science as the weighted average of the embedding vectors for each record of the respective subject by considering the weight for a given record as the inverse of its total number of subjects. For a thorough evaluation, we must also consider that many of the subjects from Web of Sciences are effectively supersets and/or intersections of other subjects, for example, those including the word \textit{multidisciplinary}. Observed are properties of the subjects, such as their spread among the embedding space, outliers, i.e., points of a certain subject that lie far away from its center, or pairwise distances between subject centers. \\
Secondly, we compare the pairwise distances in the embedding space with graph-based distances;
Here, we chose shortest path. The assumption is that the two metrics are positively correlated, stronger for close pairs and diminishing for pairs with larger distances in either metric.\\
Further, we considered the option of a dimensionality reduction of the embedding vectors - note that this is not the same as simply choosing an embedding model with fewer dimensions. We compute the explained variance of the principal component analysis (PCA) with varying dimensions.\\
The above methods are concerned with citation networks. Nevertheless, we emphasize that the approach is generalizable and can be applied to any given dataset that is comprised of textual data and also meaningfully induces a graph structure through its additional features.

\section{Computational Experiments / Study}

\subsection{Data and Setup}
The dataset comprises the Web of Science collection \cite{web-of-science} up to 2024, containing 56,143,951 records linked by 988,359,820 citations. For this study, we construct a random subsample of 48076 records by selecting each record with a probability of $10^{-3}$. 6175 ($12.8\%$) records are removed in preprocessing, leaving 41901.
Random sampling combines the advantages of a representation of the entire dataset and the feasibility of computations regarding their space and runtime requirements. \\
During pre-processing, we remove all records that have no publication year, no authors, no references, no journal, title, or unique identifier. In many cases, these fields are
empty due to missing data, and some of the above fields are needed in computations of subsequent steps. Finally, we exclude all records for which the abstract does not exceed
100 characters. This consequently leaves out those that contain invalid or empty strings or single annotations such as \textit{``A correction to this paper has been published"}.\\
We conduct all experiments on a NVIDIA Grace Hopper. Although it is entirely feasible for an architecture such as the one laid out in this paper to be implemented on commodity hardware, it does not necessarily require high-performance computing. After all, these methods are meant for data analysis and research, and as such, they are supposed to be repeatable and run with low to moderate requirements and time.

\subsection{Discussion}
\begin{figure}[h]
\includegraphics[width=0.5\textwidth]{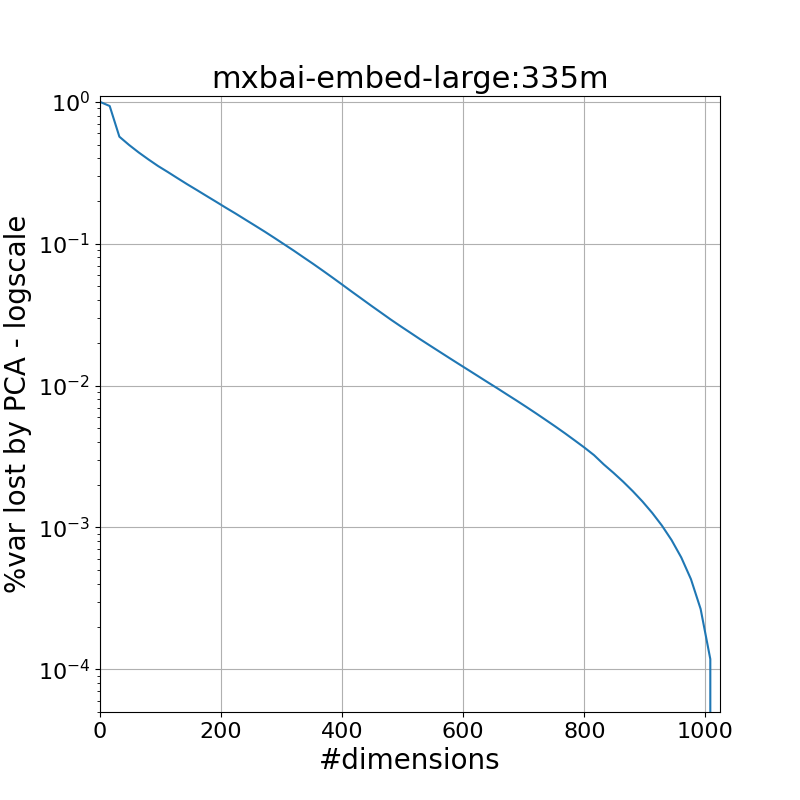}
\hfill
\includegraphics[width=0.5\textwidth]{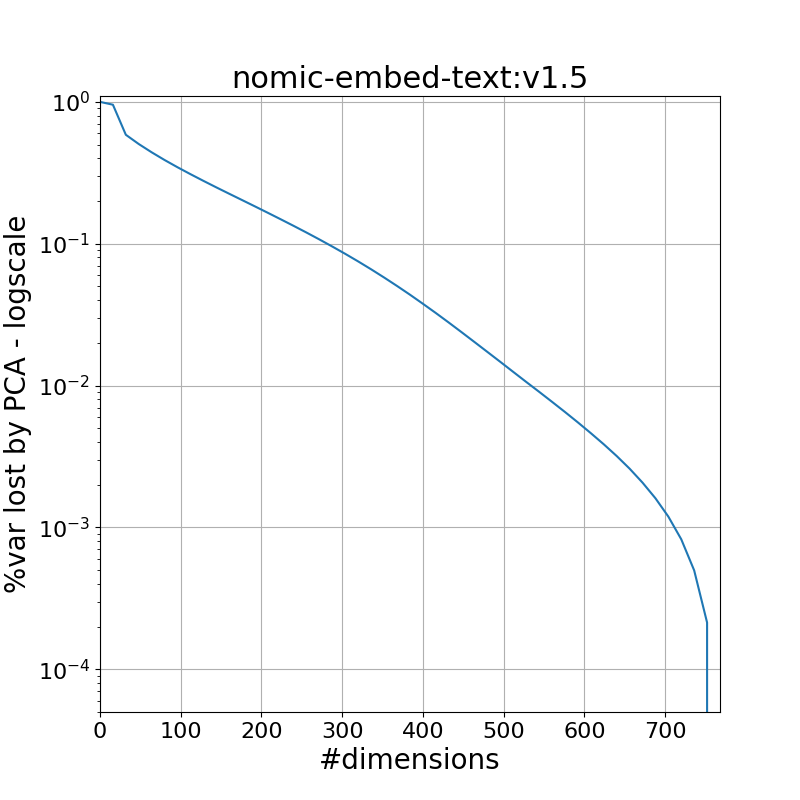}
\caption{PCA variance explained as a function of number of dimensions}
\label{fig:1}       
\end{figure}

\begin{figure}[h]
\includegraphics[width=0.5\textwidth]{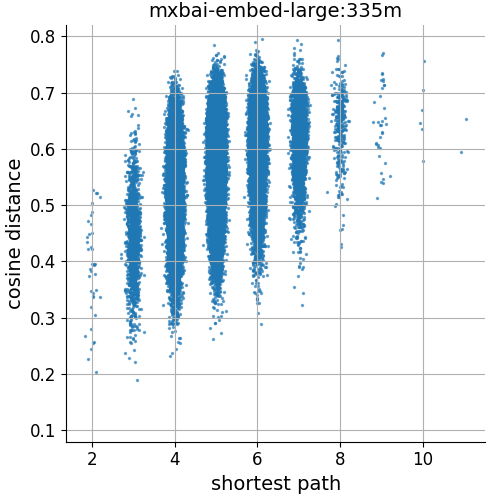}
\hfill
\includegraphics[width=0.5\textwidth]{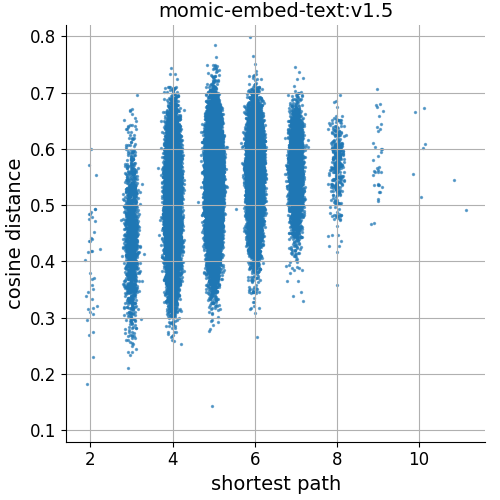}
\caption{Comparing pairwise distances of records in the embedding space and on the citation graph}
\label{fig:2}       
\end{figure}

\begin{figure}[b]
\includegraphics[width=1.35\textwidth, angle=90]{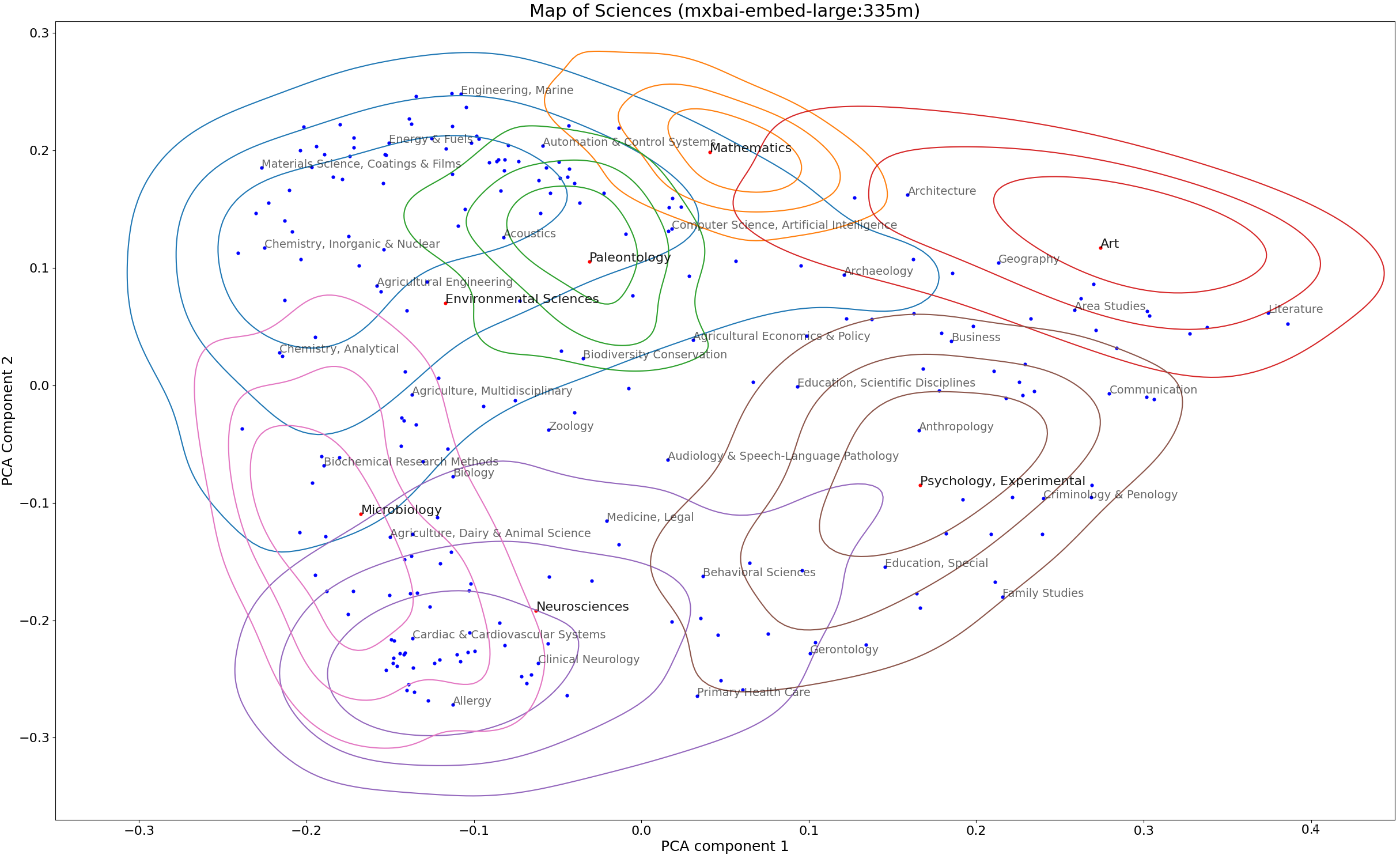}
\caption{Visualization of the embedded abstracts using PCA -  A Map of Sciences}
\label{fig:3}       
\end{figure}

A PCA analysis shows that regardless of the homogeneous text data, it seems that the embedding space of both models is populated in all dimensions (Fig. \ref{fig:1}). We conclude that dimensionality reduction as a mean to increase the performance of subsequent computations, such as clustering algorithms, may come at the cost of granularity.
\\
We randomly sample 100,000 pairs from the 48,076 records in our dataset to compute graph-based distances. We observe a positive correlation between the distances measured in the embedding space and those on the citation graph, with a PCC of $0.455$ and $0.337$ for \textit{mxbai-embed-large:335m} and \textit{nomic-embed-text:v1.5} respectively (Fig. \ref{fig:2}). 
\\
The results suggest that we can construct a combined measure of embedding distance and graph-based distance to capture both interdisciplinary and similarity. We identify an interdisciplinary set of records by a large variance in the embedding space, while the graph-based distances are small. Similarity of distant topics is the inverse; it is expressed by a proximity in the embedding space, while the graph-based distances are large. A combined measure also promises improved results on topic classification of individual papers and clustering of the entire dataset. Individually, the two approaches are already proven to be feasible \cite{clustering} \cite{emb-clustering}. Together, they have the potential to compensate each other's shortcomings in that they are methodologically different, and thereby their outliers are ideally independent. As a result, a combination of the two in the form of feature selection promises higher accuracy, which we plan to investigate in future work.\\
Computing the center-points of all subjects and observing the distribution of the records of each subject across the embedding space yields a map of sciences (Fig. \ref{fig:3}). Each dot represents one subject center. Only selected dots are labeled for increased readability. For a selection of larger subjects, a KDE (kernel density estimate) plot shows their distribution across the embedding space. The three levels correspond to 25\%, 50\%, and 75\% of all records in that subject, respectively. Within the embedding space, natural sciences, social sciences, and humanities all form larger interconnected clouds, and so do their respective branches. 
We observe varying distributions for different subjects, depending on their interdisciplinary nature. Due to multi-labeling, the human-made distinction and the observed distributions, it is evident that a classification of science can only be soft, i.e., by assigning a probability distribution over multiple subjects.

\subsubsection*{Acknowledgements}
This work is co-funded by the European Union (European Regional Development Fund EFRE, Fund No. STIIV-001). We acknowledge the use of WoS through the Kompetenznetzwerk Bibliometrie. Supported via the German Competence Network for Bibliometrics funded by the Federal Ministry of Education and Research (Grant: 16WIK2101A).
%The research for this article was conducted at the Research Campus MODAL,
%funded by the German Federal Ministry of Education and Research (BMBF)
%(Grant No. 05M14ZAM, 05M20ZBM, 05M2025)
%%%%%%%%%%%%%%%%%%%%%%%% referenc.tex %%%%%%%%%%%%%%%%%%%%%%%%%%%%%%
% sample references
% %
% Use this file as a template for your own input.
%
%%%%%%%%%%%%%%%%%%%%%%%% Springer-Verlag %%%%%%%%%%%%%%%%%%%%%%%%%%
%
% BibTeX users please use
% \bibliographystyle{}
% \bibliography{}
%

\end{document}